\documentclass[conference]{IEEEtran}

\IEEEoverridecommandlockouts

\usepackage{amsmath,amssymb,amsfonts}
\usepackage{algorithmic}
\usepackage{graphicx}
\usepackage{textcomp}
\usepackage{xcolor}
\usepackage{glossaries}
\usepackage{dirtytalk}
\usepackage{tabularray}
\usepackage{stmaryrd}
\usepackage{subfigure}
\usepackage{caption}
\usepackage{graphicx}
\usepackage{hyperref}

\usepackage[backend=biber,style=ieee]{biblatex}
\def\BibTeX{{\rm B\kern-.05em{\sc i\kern-.025em b}\kern-.08em
    T\kern-.1667em\lower.7ex\hbox{E}\kern-.125emX}}

\newcommand{\fmap}{x}
\newcommand{\sfmap}{X}
\newcommand{\sfmapVec}{\sfmap_{: \, h w}}
\newcommand{\sfmapSet}{\boldsymbol{\sfmap}}

\newcommand{\sfmapSetTrain}{\sfmapSet_{\text{train}}}
\newcommand{\sfmapSetTrainFlat}{\sfmapSet_{\text{train}}^{\text{\textit{flat}}}}
\newcommand{\sfmapSetTest}{\sfmapSet_{\text{test}}}

\newcommand{\swmap}{Y}
\newcommand{\swmapVec}{\swmap_{: \, h w}}
\newcommand{\swmapSet}{\boldsymbol{\swmap}}

\newcommand{\swmapSetTrain}{\swmapSet_{\text{train}}}

\newcommand{\sqwmap}{y^2}
\newcommand{\sqswmap}{Y^2}
\newcommand{\sqswmapVec}{\left( \sqswmap \right)_{: \, h w}}
\newcommand{\sqswmapSet}{\boldsymbol{\swmap}^2}
\newcommand{\sqswmapSetTrain}{\sqswmapSet_{\text{train}}}

\newcommand{\intrange}[1]{\llbracket 1, #1 \rrbracket}

\newcommand{\mdist}[1]{M\left( #1 \right)}
\newcommand{\mdistTwo}[1]{M^2\left( #1 \right)}

\newcommand{\euclideanNorm}[1]{\lVert \, #1 \, \rVert_2}
\newcommand{\normOne}[1]{\lVert \, #1 \, \rVert_1}

\newcommand{\whitening}[1]{\PrecMatSqrt \left( #1 - \mean \right)}

\newcommand{\mean}{\hat{\mu}}
\newcommand{\CovMat}{\hat{\Sigma}}
\newcommand{\PrecMat}{\CovMat^{-1}}
\newcommand{\PrecMatSqrt}{\left( \PrecMat \right)^{\frac{1}{2}}}
\newcommand{\R}{\mathbb{R}}
\newcommand{\Rplus}{\R_{+}}

\newglossaryentry{mvg}{
    name={MVG},
    first={multivariate Gaussian (MVG)},
    description={}
}

\newglossaryentry{cnn}{
    name={CNN},
    first={convolutional neural network (CNN)},
    description={}
}

\newglossaryentry{md}{
    name={M. distance},
    first={Mahalanobis distance (M. distance)},
    description={}
}

\newglossaryentry{ad}{
    name={AD},
    first={anomaly detection (AD)},
    description={}
}

\newglossaryentry{auroc}{
    name={AUROC},
    first={Area Under the Receiver Operating Characteristic curve (AUROC)},
    description={}
}

\newglossaryentry{aupr}{
    name={AUPR},
    first={Area Under the Precision-Recall curve (AUPR)},
    description={}
}

\newglossaryentry{aupro}{
    name={AUPRO},
    first={Area Under the Per-Region Overlap curve (AUPRO)},
    description={}
}

\newglossaryentry{resnet18}{
    name={ResNet-18},
    description={},
}

\newglossaryentry{wideresnet50}{
    name={Wide-ResNet-50-2},
    description={},
}

\newglossaryentry{effnet}{
    name={EfficientNet},
    description={},
}

\newglossaryentry{mvtecad}{
    name={MVTec-AD},
    first={MVTec Anomaly Detection (MVTec-AD)},
    description={},
}

\newglossaryentry{imgnet}{
    name={ImageNet},
    description={},
}

\newglossaryentry{opencv}{
    name={OpenCV},
    description={},
}

\newglossaryentry{padim}{
    name={PaDiM},
    first={Patch Distribution Modeling Framework (PaDiM)},
    description={},
}

\newglossaryentry{patchcore}{
    name={PatchCore},
    description={},
}

\newglossaryentry{cflowad}{
    name={CFLOW-AD},
    first={Conditional Normalizing Flow Anomaly Detection (CFLOW-AD)},
    description={},
}

\newglossaryentry{fastflow}{
    name={FastFlow},
    description={},
}

\newglossaryentry{spade}{
    name={SPADE},
    first={Semantic Pyramid Anomaly Detection (SPADE)},
    description={},
}

\newglossaryentry{gaussianad}{
    name={Gaussian-AD},
    description={},
}

\newglossaryentry{semiorthogonal}{
    name={Semi-orthogonal},
    description={},
}

\newglossaryentry{patchsvdd}{
    name={Patch SVDD},
    first={Patch-level Support Vector Data Description (Patch SVDD)},
    description={},
}

\newglossaryentry{differnet}{
    name={DifferNet},
    description={},
}

\newglossaryentry{csflow}{
    name={CSFLow},
    description={},
}

\begin{document}

\title{Visualization for Multivariate Gaussian \\ Anomaly Detection in Images}

\author{
\IEEEauthorblockN{João P C Bertoldo\textsuperscript{*}}
\IEEEauthorblockA{\textit{Centre for Mathematical Morphology (CMM)} \\
\textit{Mines Paris - PSL University}\\
Fontainebleau, France \\
jpcbertoldo@minesparis.psl.eu}
\and
\IEEEauthorblockN{David Arrustico\textsuperscript{*}}
\IEEEauthorblockA{\textit{Centre for Mathematical Morphology (CMM)} \\
\textit{Mines Paris - PSL University}\\
Fontainebleau, France \\
david-sergio.arrustico\_villanueva@etu.minesparis.psl.eu}
\thanks{\textsuperscript{*}These authors equally contributed to this work.}
}


\maketitle

\IEEEpubidadjcol


\begin{abstract}
This paper introduces a simplified variation of the PaDiM (Pixel-Wise Anomaly Detection through Instance Modeling) method for anomaly detection in images, fitting a single multivariate Gaussian (MVG) distribution to the feature vectors extracted from a backbone convolutional neural network (CNN) and using their Mahalanobis distance as the anomaly score. 
We introduce an intermediate step in this framework by applying a whitening transformation to the feature vectors, which enables the generation of heatmaps capable of visually explaining the features learned by the MVG.
The proposed technique is evaluated on the MVTec-AD dataset, and the results show the importance of visual model validation, providing insights into issues in this framework that were otherwise invisible. 
The visualizations generated for this paper are publicly available at \href{https://doi.org/10.5281/zenodo.7937978}{\nolinkurl{https://doi.org/10.5281/zenodo.7937978}}.
\end{abstract}


\section{Introduction} 

\Gls{ad} consists of identifying samples that deviate from a semantic concept \cite{chandola2009anomaly}. 
It is important in many fields because it allows one to flag unexpected data points, which may represent errors, malfunctions, fraud, etc. 
By detecting anomalies, one can identify and mitigate potential problems before they cause significant damage, and gain insights into the underlying data that might not be apparent otherwise. 
In image \gls{ad}, these findings can be done at different scales, most commonly at the pixel and image levels. 

In recent years, deep learning techniques for \gls{ad} tasks have gained considerable attention \cite{ruff2021unifying}, which can be attributed to their ability to generate meaningful data representations in the form of feature maps.
In this line, ResNet \cite{he_deep_2015} marked a notable breakthrough in the field of deep neural networks, for instance, by achieving success on \gls{imgnet} tasks thanks to the use of residual representations and shortcut connections.

In response to the growing demand for publicly available image datasets for \gls{ad} research, \cite{bergmann_mvtec_2019} introduced the \gls{mvtecad} dataset, which comprises over 5000 images across 15 categories. 
Anomalies in this dataset are understood as defects on textures and objects (i.e. \say{normal} images consist of defectless samples), and their masks (pixel-wise labels) are provided at high resolution.

Previous works have exploited the use of \gls{mvg} distributions and the \gls{md} to detect anomalies in \gls{mvtecad} in an unsupervised way. 
The framework proposed in \cite{rippel2021gaussian} first attacked the problem at the image level and posterior works such as \cite{defard2021padim} have adapted their approach to the pixel level.
Despite their reasonable performances, these papers lack visual explanations capable of validating or debugging their models, which we aim to cover in this paper.

In a related vein, the field of neural network visualization has also witnessed significant breakthroughs. 
For instance, \cite{bau2020understanding} successfully identified units within image classifiers that corresponded to human-interpretable concepts. 
Inspired by their visualization strategy, the present paper aims to adapt their approach to Gaussian \gls{ad} models.


\section{Method}\label{sec:method}

We propose a simplified variation of \gls{padim} \cite{defard2021padim} which, instead of fitting one Gaussian distribution per pixel position, fits all pixel-wise feature vectors under a single distribution.

\subsection{Feature maps}

Let $\fmap \in \R^{H \times W}$ be a feature map extracted from a backbone \gls{cnn}, where $H$ and $W$ are, respectively, its height and width -- notice that they are generally smaller than the image's height and width. 
The collection of feature maps (stack operation) from a single layer of the backbone \gls{cnn} is denoted $\sfmap \in \R^{C \times H \times W}$, where $C$ is its number of channels (i.e. different feature maps). 
A collection of feature maps from $B$ images is denoted, in bold, $\sfmapSet \in \R^{B \times C \times H \times W}$, and we add subscripts to differentiate the train set ($\sfmapSetTrain$; only normal images) and the test set ($\sfmapSetTest$; normal and anomalous images).

We use the term \say{[pixel-wise] feature vector} and denote it as $\sfmapVec$ to refer to the vector along the channel dimension ($:$ denotes \say{all values}) at height position $h \in \intrange{H}$ and width position $w \in \intrange{W}$ of the feature maps $\sfmap$, where $\intrange{I}$ denotes the set of integer values $ \{ 1 , \dots , I \}$.

\subsection{Whitened feature maps}

We fit a \glsfirst{mvg} distribution with the feature vectors in $\sfmapSetTrain$ with all pixel positions confounded. 
A flattening operation is done on $\sfmapSetTrain$ so that all dimensions except the channel dimension are reduced.
Thus, we obtain $\sfmapSetTrainFlat \in \R^{BHW \times C}$, from which the empirical mean vector $\mean \in \R^{C}$ and the unbiased empirical covariance matrix $\CovMat \in \R^{C \times C}$ are computed, respectively, by

\begin{equation} \label{eq:mean_vector}
    \mean = \frac{1}{BHW} \sum_{i=1}^{BHW} \left( \sfmapSetTrainFlat \right)_{i \, :}
\end{equation}

\noindent
and

\begin{equation} \label{eq:cov_matrix}
    \CovMat = \frac{1}{BHW - 1}\sum_{i=1}^{BHW} \left[ \left( \sfmapSetTrainFlat \right)_{i \, :} - \mean \right] \left[ \left( \sfmapSetTrainFlat \right)_{i \, :} - \mean \right]^T
    \; .
\end{equation}

\noindent
At inference time, the \glsfirst{md} of a feature vector $\sfmapVec$

\begin{equation} \label{eq:ref_MD}
    \mdist{\sfmapVec} = \sqrt{ \left(\sfmapVec - \mean \right)^T \CovMat^{-1} \left(\sfmapVec - \mean \right) }
\end{equation}

\noindent
is assigned to each pixel in the feature maps resolution as its anomaly score (higher means \say{more anomalous}).
However, Eq.~\ref{eq:ref_MD} can be rewritten as 

\begin{equation} \label{eq:ref_MD2}
\begin{split}
&\mdistTwo{\sfmapVec}
  = \left(\sfmapVec-\mean \right)^T \PrecMat \left(\sfmapVec-\mean \right) \\
 & = \left(\sfmapVec-\mean \right)^T \PrecMatSqrt \PrecMatSqrt \left(\sfmapVec-\mean \right) \\
 & = \left[ \PrecMatSqrt \left(\sfmapVec-\mean \right) \right]^T \left[ \PrecMatSqrt \left(\sfmapVec-\mean \right) \right] \\
 & = \left[ \swmapVec \right]^T \; \swmapVec  
   = \euclideanNorm{\swmapVec}^2
 \; ,
\end{split}
\end{equation}

\noindent
where $\PrecMatSqrt$ is the square root matrix of $\PrecMat$ (\textit{not} the point-wise square root operator) and 

\begin{equation} \label{eq:whitening}
\swmapVec = \whitening{\sfmapVec}
\end{equation}

\noindent
is the result of an empirical whitening transformation\footnote{Eq.~\ref{eq:whitening} is called \say{whitening} transformation because the covariance matrix of its image on the training data is the identity matrix, therefore a white noise. \say{Empirical} refers to using the empirical parameters $\mean$ and $\CovMat$.} applied to $\sfmapVec$, therefore we name it a \textit{whitened} feature vector.
Analogously, $\swmap \in \R^{C \times H \times W}$ is referred to as the \textit{whitened} feature maps, and its channels are referred to as \say{components} as a reminder that they correspond to the projection of the (centered) feature vector on each \textit{eigencomponent} of $\PrecMat$.

\subsection{Squared whitened feature maps}

As shown in Eq.~\ref{eq:ref_MD2}, the \gls{md} of $\sfmapVec$ is equivalent to Euclidean norm of $\swmapVec$. 
Since $\cdot^2$ is monotonic, the squared \gls{md} $\mdistTwo{\cdot}$ has the same ordering than $\mdist{\cdot}$. 
Thus the anomaly score of $\sfmapVec$ can be explained as the sum of the components of the \textit{squared} whitened feature vector $\sqswmapVec^2$, where $\cdot^2$ is a point-wise operator, as

\begin{equation} \label{eq:ref_MD3}
\begin{split}
\mdistTwo{\sfmapVec} &= \euclideanNorm{\swmapVec}^2 \\ 
    &= \normOne{\sqswmapVec} = \sum_{c = 1}^{C} \left( \sqswmap \right)_{chw}
 \; .
\end{split}
\end{equation}




\section{Experiments}

We leverage the equivalence proposed in Eq.~\ref{eq:ref_MD3} to generate visual explanations of the anomaly score maps $\mdistTwo{\sfmap} \in \left( \Rplus \right)^{H \times W}$, where 

\begin{equation}
    \left( \mdistTwo{\sfmap} \right)_{hw} = \mdistTwo{\sfmapVec} \, \forall \, (h, w) \in \intrange{H} \times \intrange{W} 
    \; .
\end{equation}


\subsection{Backbone}\label{sec:backbone}

We use a \gls{resnet18} \cite{he_deep_2015} pre-trained on the \gls{imgnet} dataset (\texttt{torchvision} version \texttt{0.15.2}, weights \texttt{ResNet18\_Weights.IMAGENET1K\_V1}) as the backbone \gls{cnn} to extract feature maps.
The feature maps are extracted from its four major blocks: \say{layer1}, \say{layer2}, \say{layer3}, and \say{layer4}.
The figures with their visualizations (Sec.~\ref{sec:results-and-discussion}) indicate their anomaly segmentation performances on the test set in terms of pixel-wise \gls{auroc}, pixel-wise \gls{aupr}, and \gls{aupro}~\cite{bergmann_mvtec_2019}.

\subsection{Data}

The model proposed in Eq.~\ref{eq:mean_vector} and Eq.~\ref{eq:cov_matrix} is independently trained on each layer of the backbone \gls{cnn} and each category of the \gls{mvtecad} dataset \cite{bergmann_mvtec_2019}.
Then, the proposed whitening transformation (Eq.~\ref{eq:whitening}) is applied on the feature maps.
Finally, we generate the \textit{squared} whitened feature maps $\sqswmap$ (point-wise square of $\swmap$).

All the visualizations are publicly available on Zenodo at \href{https://doi.org/10.5281/zenodo.7937978}{\nolinkurl{https://doi.org/10.5281/zenodo.7937978}} \cite{bertoldo_joao_p_c_pixel-wise_2023} 
However, this paper is limited to the category \say{hazelnut} for the sake of space and simplicity. 
Similar observations were seen in other categories.

\subsection{Visualizations}

All visualizations consist of a superposition of two images using alpha blending with $\alpha = 0.5$: the input image and a \emph{heatmap}, which can be a \textit{squared component map} $\sqwmap = \left( \sqswmap \right)_{c::}$ of a given component $c$ or the squared anomaly score map $\mdistTwo{\sfmap}$.
The heatmaps are upsampled to $224 \times 224$ to match the input image resolution\footnote{$\mdistTwo{\sfmap}$ is first computed in its own (lower) resolution and then resized.}.
We use \gls{opencv} (Python library version \texttt{4.7.0.68}) with bilinear interpolation and the plots are generated with \texttt{matplotlib} (version \texttt{3.6.3}). 

\subsection{Color scale}\label{sec:color-scale}

The heatmaps use the color scale \say{inferno} with the minimum value set to zero and the maximum value set to a per-plot statistic.
In simple terms, we use percentile values to avoid losing color resolution when extreme values cause the maximum to explode.
For the heatmaps of $\mdistTwo{\sfmap}$, the maximum value is set to the $99$th percentile of all values of $\mdistTwo{\sfmap}$ in the dataset split (train or test) being visualized.
For the heatmaps of $\sqwmap$, we propose two color scaling strategies, which serve different purposes, referred to as \say{per-component color scale} and  \say{cross-component color scale}.

\paragraph*{Per-component color scale}
a different maximum value is chosen for each component independently; the color scale's maximum value of a fixed component $c$ is the $99$th percentile of all values of $\left( \sqswmapSet \right)_{:c::}$ in the dataset split (train or test) being visualized.
This strategy is useful to understand the information encoded in each component because it provides better visibility of the nuances across different images.

\paragraph*{Cross-component color scale}
a single maximum value is chosen for all components; the color scale's maximum value is the $99$th percentile of all values of $\sqswmapSet$ in the dataset split (train or test) being visualized.
With this strategy, all the heatmaps are comparable in magnitude, so the \say{visual sum} of the heatmaps from the same image corresponds to the anomaly score map (their component-wise sum, cf. Eq.~\ref{eq:ref_MD3}).


\subsection{Component Sorting (Single-component AUROC)}\label{sec:component-sorting}

As a result of the whitening transformation\footnote{We use the method \texttt{whiten()} from \texttt{scipy.stats.Covariance}, which takes the eigendecomposition of $\CovMat$ as constructor arguments.}, the value of $\swmap_{chw}$ corresponds to the projection of the vector $\sfmapVec - \mean$ onto the eigenvector (from $\CovMat$) with the $c$-th smallest eigenvalue associated to it. 
In other words, the channels of $\swmap$ are sorted by the eigenvalues of $\CovMat$ in ascending order.
However, we found it more helpful to sort them by their individual performances. 
Each squared component $\sqwmap$ is interpreted as anomaly score map (as if $\swmap$ was reduced to a single component) and its respective (pixel-wise) \gls{auroc}, later referred to as \say{single-component \gls{auroc}}, is used to sort the components of each layer separately.

\begin{figure*}[!t]
\centering
\subfigure[layer1]{\label{fig:l1_00}\includegraphics[width=0.49\textwidth]{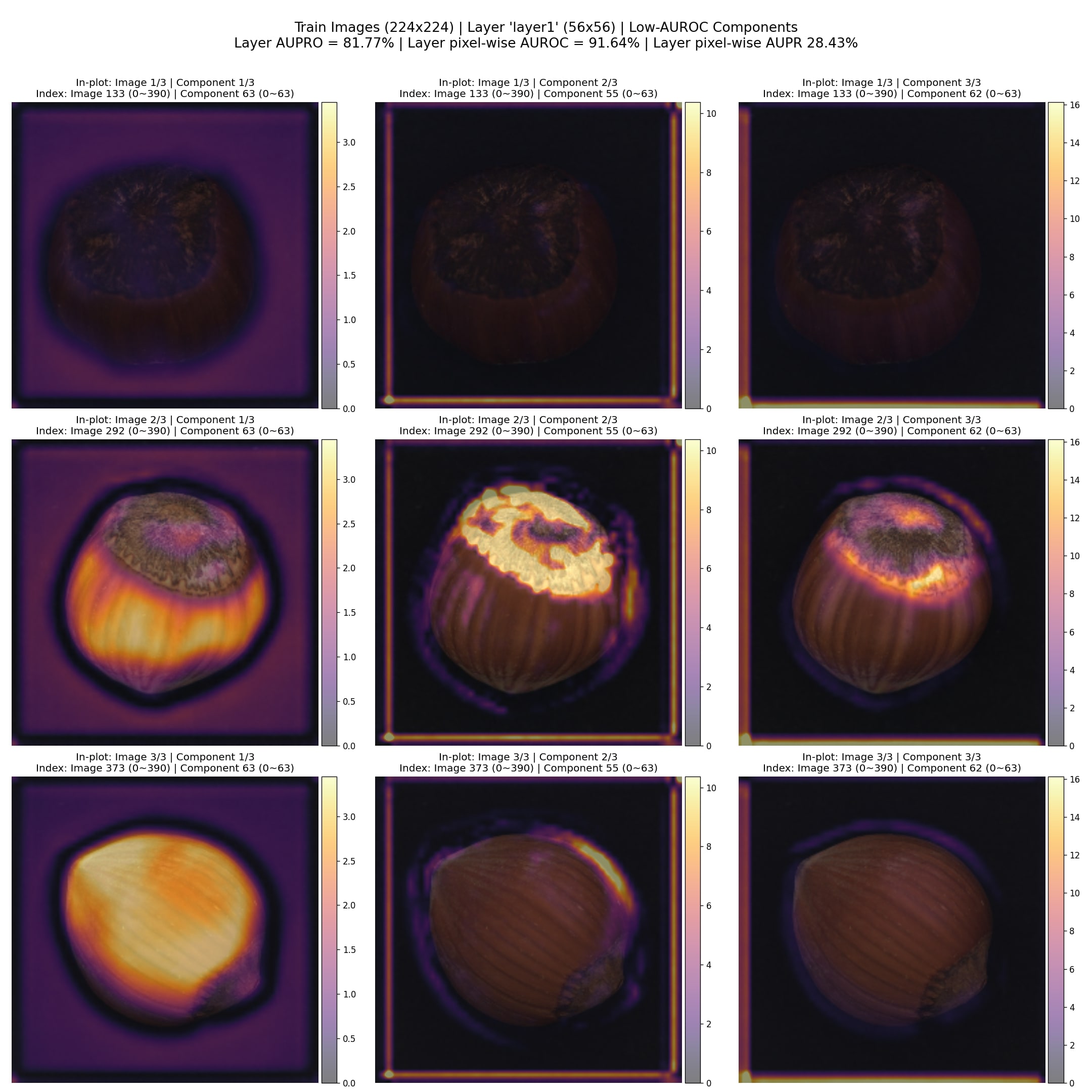}}
\hfill
\subfigure[layer2]{\label{fig:l2_00}\includegraphics[width=0.49\textwidth]{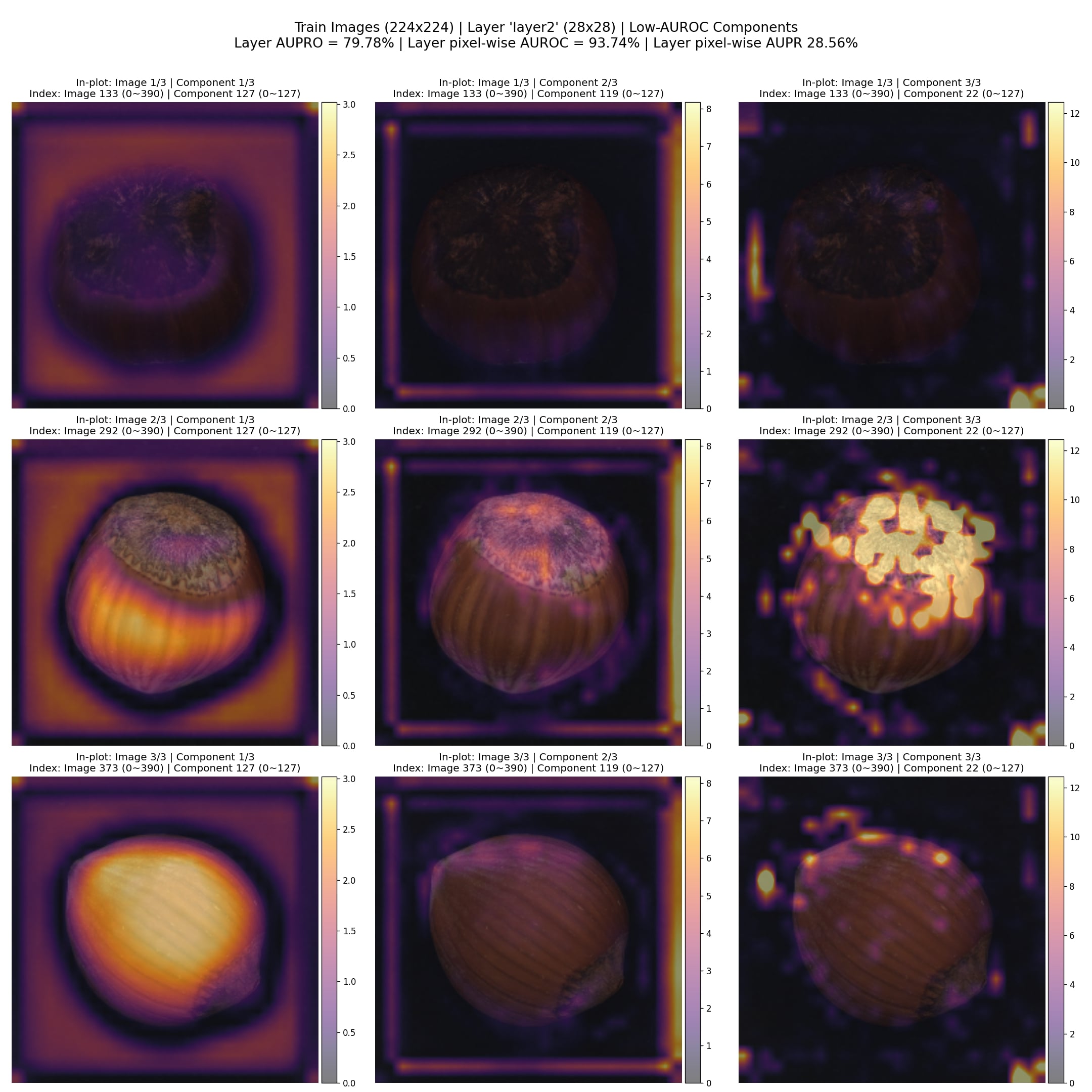}}
\\ 
\vspace{-5pt}
\subfigure[layer3]{\label{fig:l3_00}\includegraphics[width=0.49\textwidth]{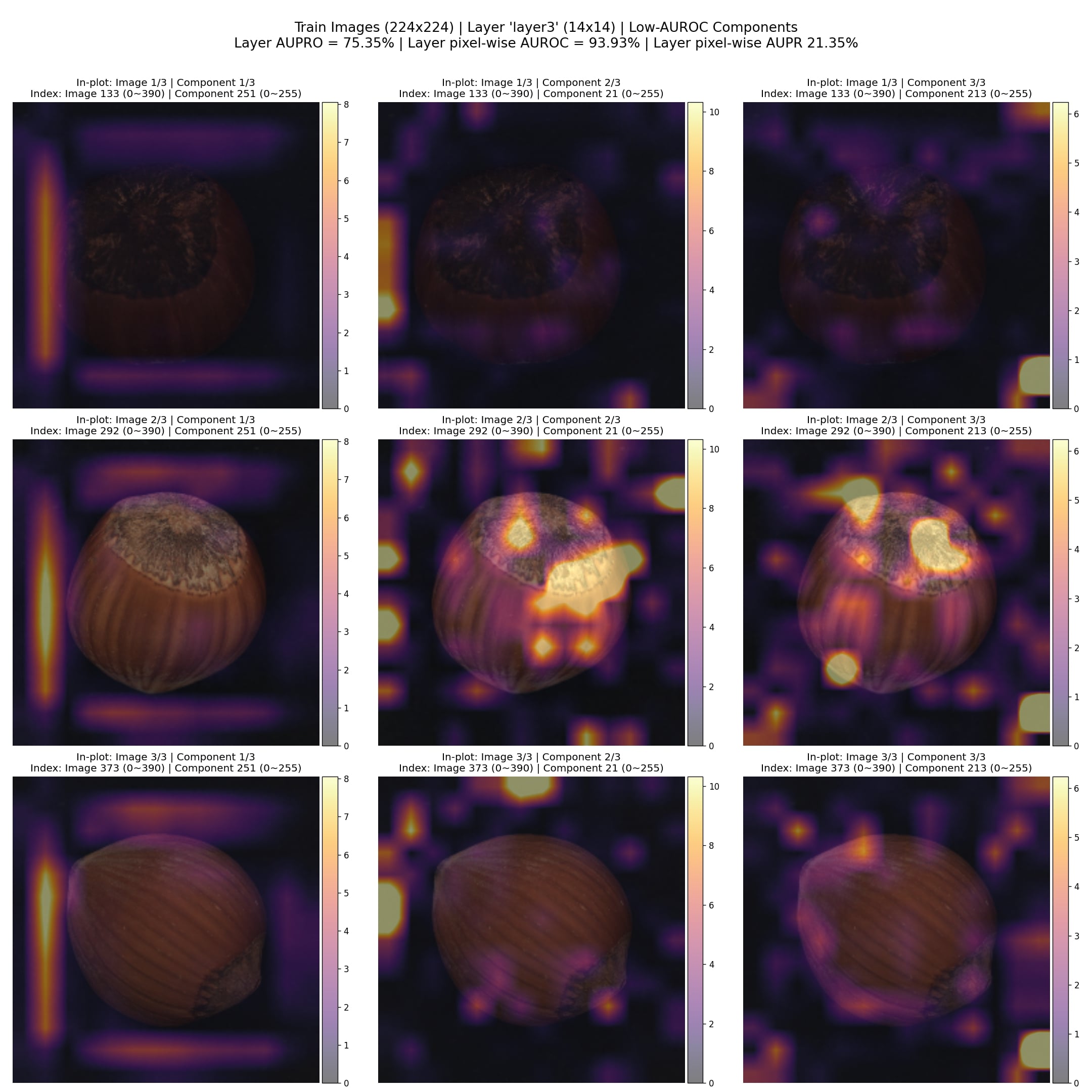}}
\hfill
\subfigure[layer4]{\label{fig:l4_00}\includegraphics[width=0.49\textwidth]{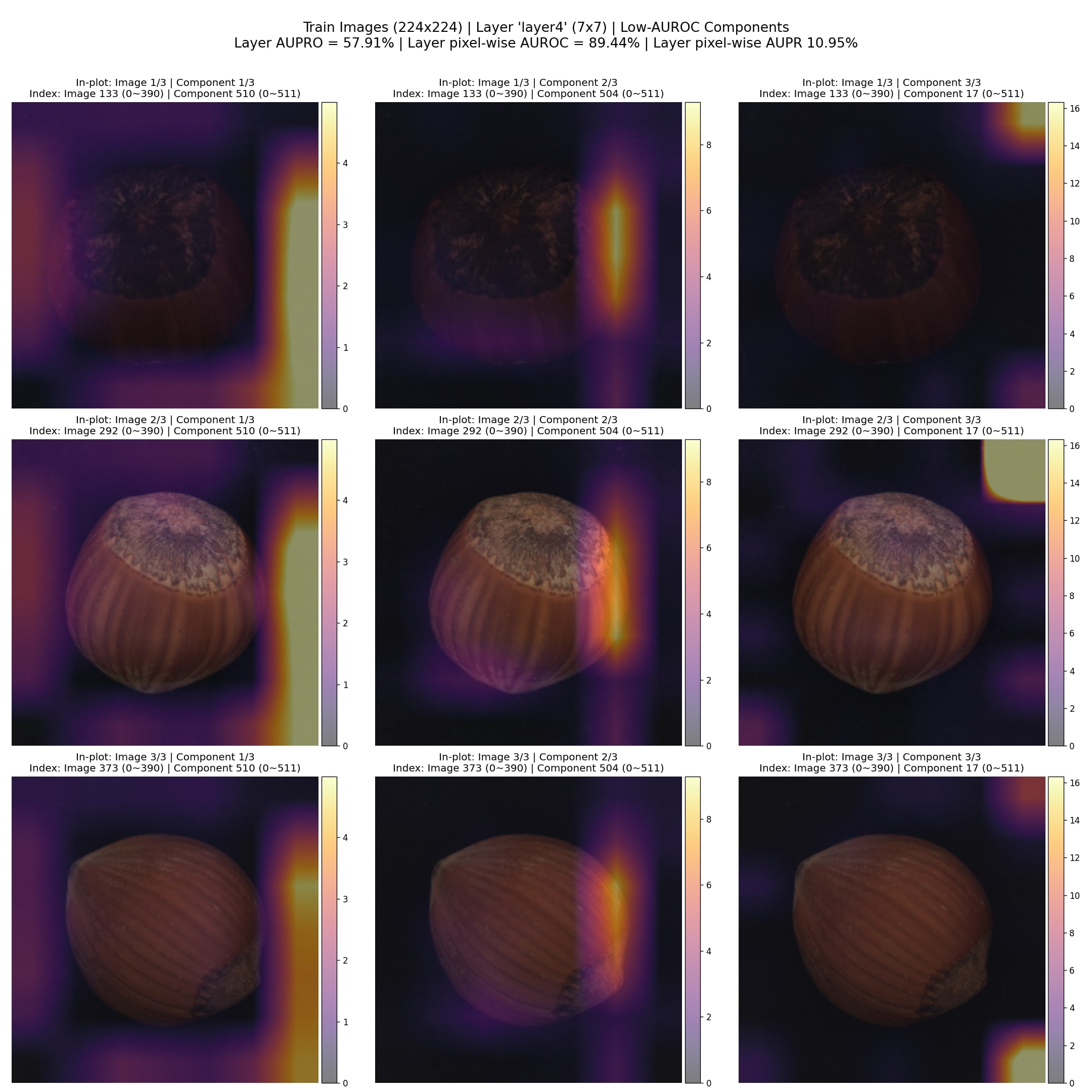}}
\vspace{-5pt}
\caption{
    Visualizations on training (normal) images. Selection of low-AUROC components.
}
\label{fig:l1l2l3l4_00}
\end{figure*}

\section{Results and Discussion}\label{sec:results-and-discussion}

Visual inspection of a large number of images revealed that some weaknesses of the model can be easily identified, but understanding how components interact and build a correct anomaly score map is not as simple. 
We present a selection of visualizations that summarize the major insights learned.

The upper title of each figure displays the segmentation performance of the model being visualized (using all the components).
Moreover, on top of each visualization, the index of the image and the index of the component in $\swmap$ are indicated. 
\say{In-plot} indices refer to the (1-starting) index in the grid of that figure. 
\say{Index} indices refer to the (0-starting) index of the image as sorted in \gls{mvtecad}, and to the (0-starting) index of the component as sorted by increasing order of eigenvalues from $\CovMat$ (cf. Sec.~\ref{sec:component-sorting}). 

First, we focus our analysis on \textit{bad} (low single-component \gls{auroc}) components, whose weaknesses can be seen even on the train set.
Then we visualize these same components on test anomalous images, which show that some artifacts are attenuated while others persist.
Finally, we show test anomalous images superposed with \textit{good} (high single-component \gls{auroc}) components and their final anomaly score map.

\begin{figure*}[!t]
\centering
\subfigure[layer1]{\label{fig:l1_01}\includegraphics[width=0.49\textwidth]{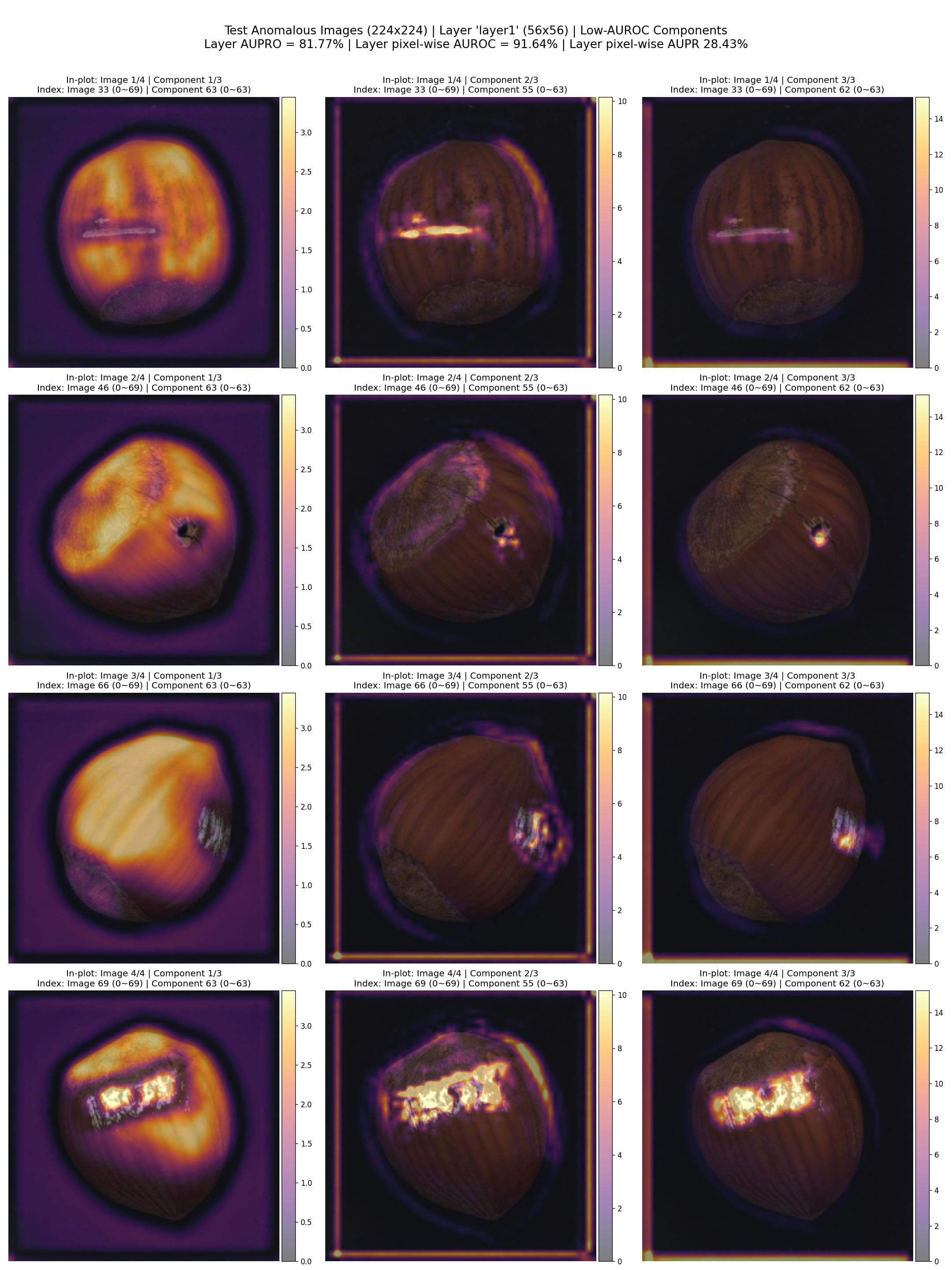}}
\hfill
\subfigure[layer2]{\label{fig:l2_01}\includegraphics[width=0.49\textwidth]{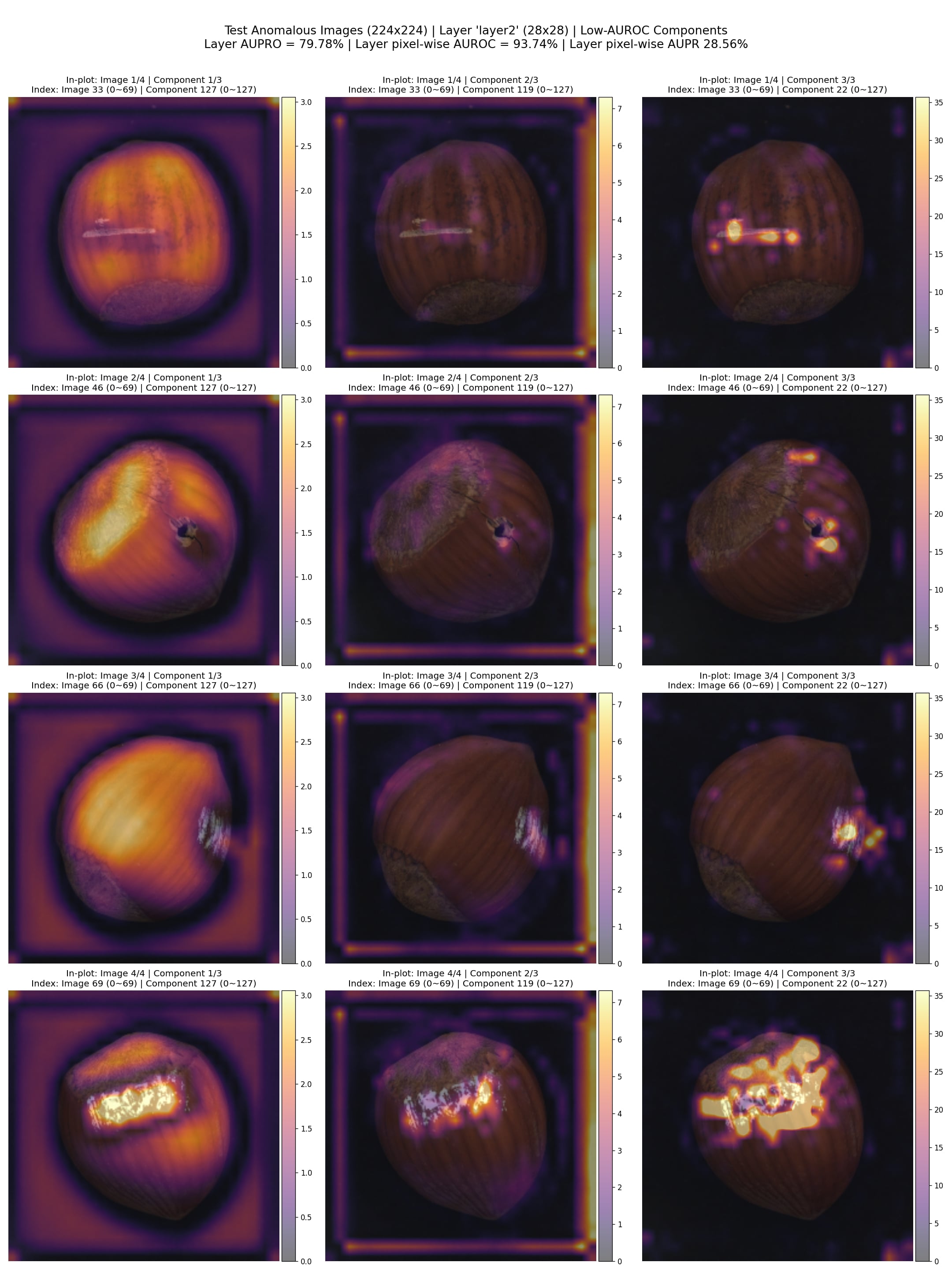}}
\vspace{-5pt}
\caption{
    Visualizations on test anomalous images. Selection of low-AUROC components.
}
\label{fig:l1l2_01}
\end{figure*}

\subsection{Low-AUROC components}

In this subsection, all heatmaps use the per-component color scale (cf Sec.~\ref{sec:color-scale}) to better visualize their artifacts.
Fig.~\ref{fig:l1_00}, Fig.~\ref{fig:l2_00}, Fig.~\ref{fig:l3_00}, and Fig.~\ref{fig:l4_00} display heatmaps of three squared components among the \textit{worst} (lowest single-component \gls{auroc}) respectively from layer1, layer2, layer3, and layer4.
They all show the same three \textit{normal images from the training set}.
Fig.~\ref{fig:l1_01} and Fig.~\ref{fig:l2_01} display, respectively, the same layers and components present in Fig.~\ref{fig:l1_00} and Fig.~\ref{fig:l2_00}, but superposed on \textit{anomalous images from the test set}.
Anomalous images from layer3 and layer4 were omitted due to the lack of space but similar phenomena have been observed.

\subsubsection{Resolution mismatch}\label{sec:resolution-mismatch}

\cite{defard2021padim} proposed that deeper layers achieve lower performances -- despite their higher number of features -- because they are too biased by the classes from the pre-training dataset, but our results show an alternative (or complementary) explanation.
Fig.~\ref{fig:l1l2l3l4_00} shows that deeper layers have an increasing problem of resolution mismatch.
As downsampling operations are chained, the feature maps' resolutions become too coarse;
layer4, for instance, is $7 \times 7$ while the input image is $224 \times 224$, thus a  pixel in the former represents a patch of $32 \times 32$ from the latter.

\subsubsection{Whitened component value ranges}\label{sec:value-range}

for any component $c$, $99.7\%$ of the values in $\left( \swmapSetTrain \right)_{:c::}$ are expected to be in the interval $[-3, 3]$ due to the whitening operation (empirical rule for a standard normal distribution), so the values in $\left( \sqswmapSetTrain \right)_{:c::}$ are expected to be in the interval [0, 9].
However, when setting the maximum color scale with the $99$th percentiles of $\left( \swmapSetTrain \right)_{:c::}$ (cf. Sec.~\ref{sec:color-scale}), inconsistent behavior is observed in some cases.
For instance, in Fig.~\ref{fig:l1l2l3l4_00}, component 62 in layer1 and component 17 in layer4 have saturation values around 16, which is higher than expected, but component 63 in layer1, component 127 in layer2, and component 510 in layer4 have saturation values around $3 \sim 5$, which is lower than expected.

\subsubsection{Border effects}\label{sec:border-effects}

several components show input-independent artifacts (i.e. they happen no matter the content of the input image) on the borders and corners of the heatmaps.
In Fig.~\ref{fig:l1l2l3l4_00}, components 55 and 62 from layer1, component 119 from layer2, component 133 from layer3, and components 510 and 133 from layer4 show artifacts along the borders of the image (vertical and horizontal strips of high activation).
Also in Fig.~\ref{fig:l1l2l3l4_00}, component 22 from layer2, component 213 from layer3, and component 17 from layer4 similarly show artifacts on the corners of the images.
We hypothesize these issues ensue from the padding used in the backbone \gls{cnn}: as multiple padded convolutions are successively applied, the border effects accumulate and the feature vectors near the borders shift to a different distribution mode, which makes them unusual (higher anomaly score) compared to the pixels inside the image, which are the majority.

\subsubsection{Some border effects are worse than others}\label{sec:some-border-effects-are-worse}

the visualizations of anomalous images in Fig.~\ref{fig:l1l2_01} show that the presence of border effects does not necessarily indicate that a component is bad for detecting anomalous pixels.
For component 33 from layer1 and component 119 from layer2, the border effects are so pronounced that even real anomalies have lower values (cf. images 33, 46, and 66).
However, component 22 in layer2 activates more intensely on the anomalies than on the corner artifacts, making them practically disappear;
notice that the upper bound of the color scale in Fig.~\ref{fig:l2_01} is higher than in Fig.~\ref{fig:l2_00} ($\approx 35$  against $\approx 12$ respectively). 

\subsubsection{Foreground-background frontier}

particularly within object categories (e.g. \say{hazelnut}), several components seem to detect the border between the object (foreground) and the background.
When this behavior is observed, in most cases, the foreground-background border shows high activation. Interestingly, the opposite also happens in a few cases such as in Fig.~\ref{fig:l1l2l3l4_00} for component 63 in layer1 and component 127 in layer2.
Notice that this phenomenon persists in anomalous images (Fig.~\ref{fig:l1l2_01}), and some anomalies even have low values compared to the normal foreground and background (e.g. images 46 and 66 in Fig.~\ref{fig:l1_01}), which contradicts the expected behavior of the model.

\begin{figure*}[!t]
\centering
\subfigure[layer1]{\label{fig:l1_02}\includegraphics[width=0.49\textwidth]{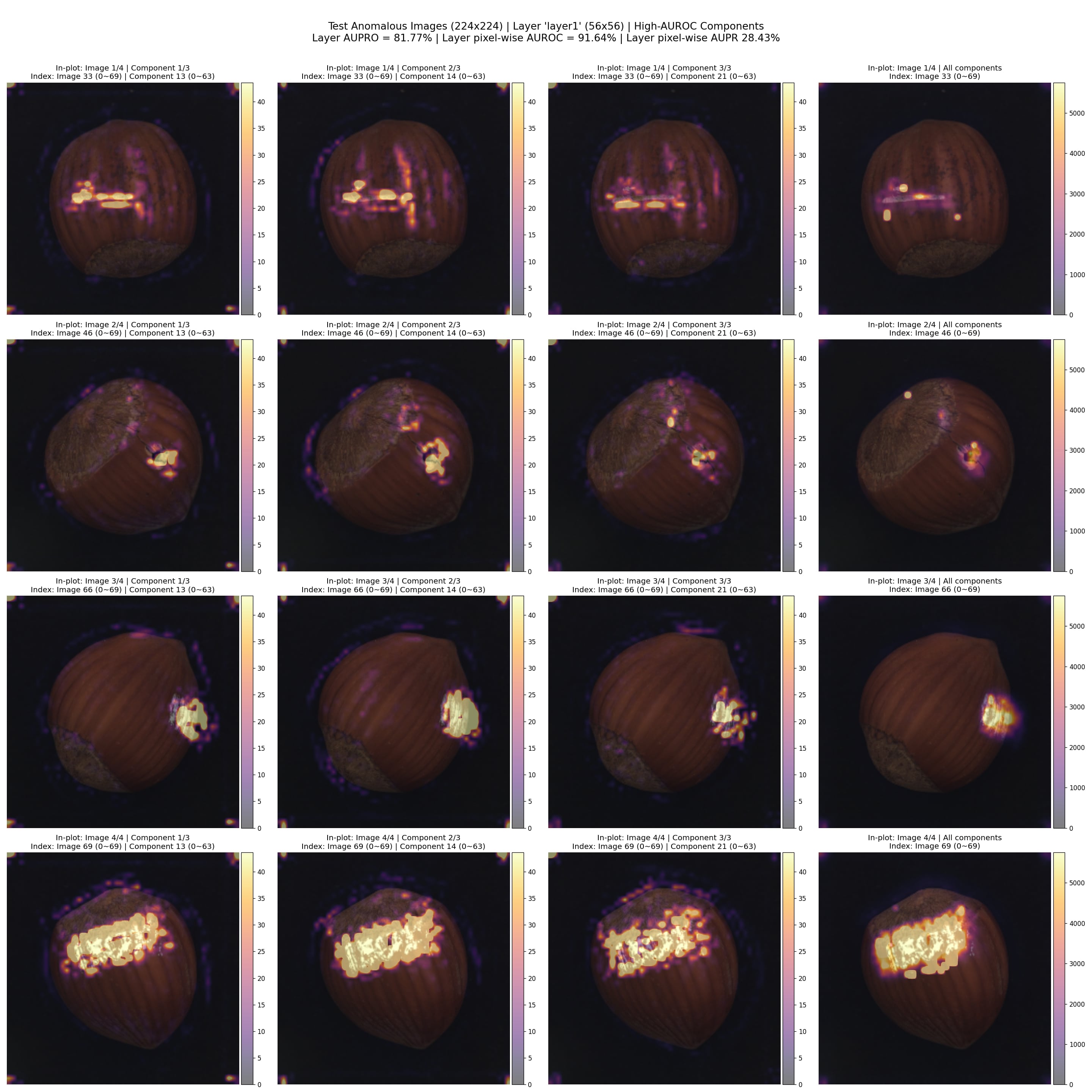}}
\hfill
\subfigure[layer2]{\label{fig:l2_02}\includegraphics[width=0.49\textwidth]{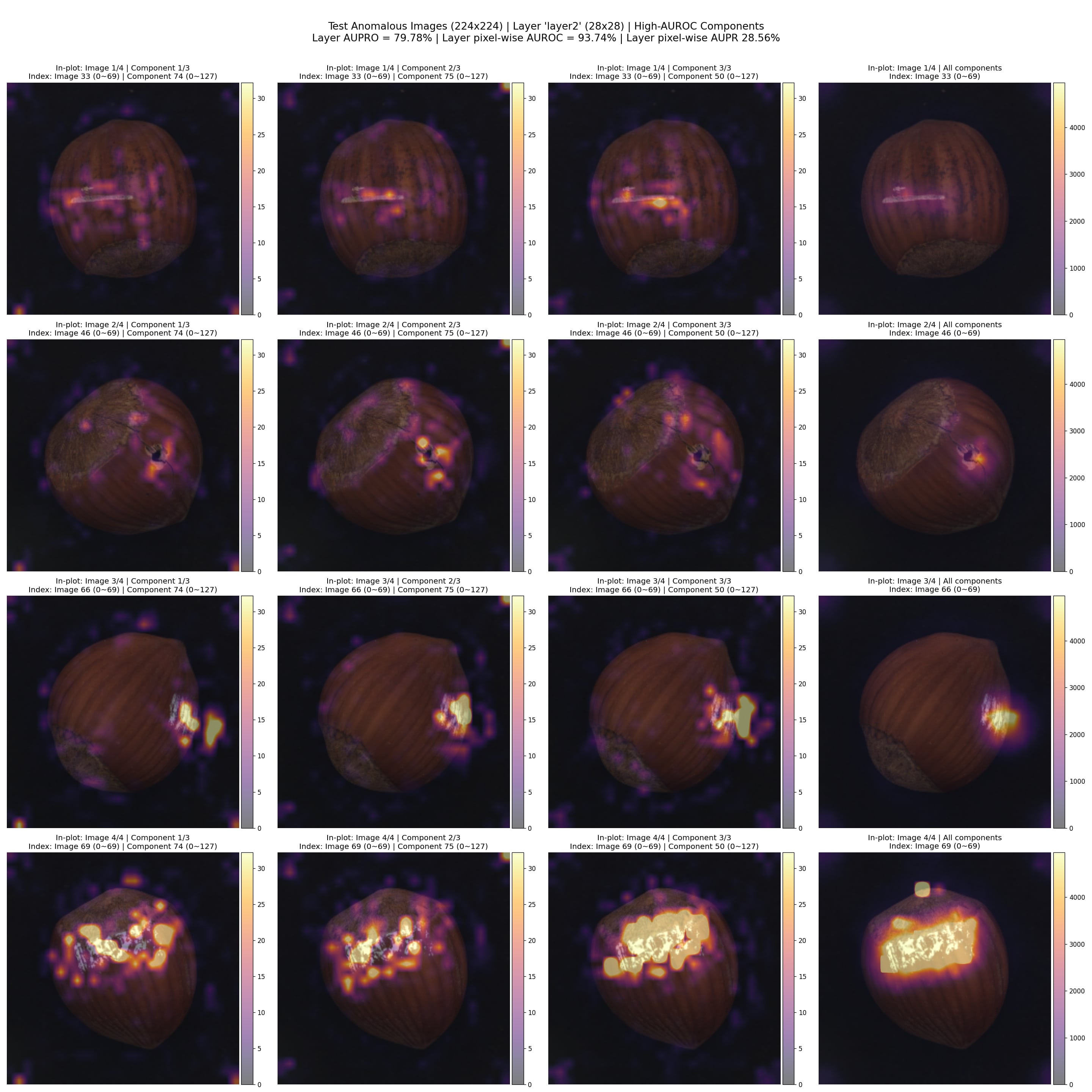}}
\vspace{-5pt}
\caption{
    Visualizations on test anomalous images. Selection of high-AUROC components.
}
\label{fig:l1l2_02}
\end{figure*}

\subsection{High-AUROC components}


In this subsection, all heatmaps use the cross-component color scale (cf. Sec.~\ref{sec:color-scale}) to better visually compare how the values of different components interact. 
Fig.~\ref{fig:l1l2_02} displays heatmaps of three components among the \textit{best} (highest single-component \gls{auroc}) and the anomaly score map (right-most heatmap at each row) from layer1 and layer2.
They both show the same four \textit{anomalous images from the test set} present in Fig.~\ref{fig:l1l2_01}.
The analogous visualizations from layer3 and layer4 are omitted due to the lack of space but similar phenomena have been observed.

The border/corner artifacts mentioned in Sec.~\ref{sec:border-effects} are also present in some cases such as components 13 and 14 in layer1 and components 74 and 75 in layer2, but they all show high values on the \textit{actual} anomalous pixels.

It can also be observed that there is some degree of \say{agreement} on the anomalous pixels, but rather \say{disagreement} on the false positive pixels.
In other words, the high-value regions on the anomalies (thus true positives) intersect across components, redundantly detecting abnormalities. 
However, the high-value regions on normal pixels (thus false positives) are randomly placed.
The anomaly score map confirms this explanation as the false positive regions are not visible because the sum of true positive values shadows them in terms of magnitude.

Finally, we observed that some types of anomaly have higher scores than others; 
for instance, compare the anomaly score maps of images 33 and 69 in Fig.~\ref{fig:l2_02}.


\section{Conclusion}


We proposed a simplified variation of \gls{padim} \cite{defard2021padim} for anomaly detection in images that fits a single \glsfirst{mvg} distribution to pixel-wise feature vectors extracted from a \gls{resnet18}. 
We showed that the \glsfirst{md} (i.e. the anomaly score) of each pixel can be expressed as the Euclidean norm of its image from an empirical whitening transformation.
Then, we explore this equivalence to generate visual explanations for our model using heatmaps of the individual axes of these \textit{whitened} feature vectors. 

The experiments conducted on the \gls{mvtecad} dataset demonstrate the utility of the proposed visualizations to provide insights into the behavior of our model, for instance revealing pixel position-related artifacts and showing that deep feature maps have bad performance because their resolution is too low.    
Our analysis suggests that the model could be improved by removing artifact-ful heatmaps (i.e. dimension reduction), adjusting the input-feature map resolution mismatch, and taking the padding effects (from the \gls{cnn} architecture) into account. 






\printbibliography

\end{document}